\documentclass{article}

\usepackage[preprint,nonatbib]{neurips_2025}

\usepackage[utf8]{inputenc} 
\usepackage[T1]{fontenc}    
\usepackage{hyperref}       
\usepackage{url}            
\usepackage{booktabs}       
\usepackage{amsfonts}       
\usepackage{nicefrac}       
\usepackage{microtype}      
\usepackage{xcolor}         
\usepackage[numbers, sort&compress]{natbib}
\usepackage{supertabular}

\usepackage{graphicx}

\usepackage{listings}
\usepackage{xcolor} 

\usepackage[utf8]{inputenc} 
\usepackage[T1]{fontenc}    
\usepackage{hyperref}       
\usepackage{url}            
\usepackage{booktabs}       
\usepackage{amsfonts}       
\usepackage{nicefrac}       
\usepackage{microtype}      
\usepackage{xcolor}         
\usepackage{tcolorbox}
\usepackage{pgfplots}
\usepackage{geometry}
\usepackage{subcaption}
\pgfplotsset{width=\textwidth,compat=1.18}
\title{CHANCERY: Evaluating Corporate Governance Reasoning Capabilities in Language Models}

\author{%
  Lucas Irwin\thanks{The first two authors contributed equally to this work.} \\
  Sentient Labs\\
  \texttt{lucas@sentient.xyz} \\
   \And
   Arda Kaz{$^\star$} \\
  Sentient Labs\\
   \texttt{arda@sentient.xyz} \\
   \AND
   Peiyao Sheng \\
   Sentient Labs\\
   \texttt{peiyao@sentient.xyz} \\
   \And
   Sewoong Oh \\
   Sentient Labs\\
   \texttt{sewoong@sentient.xyz}
   \And
   Pramod Viswanath \\
   Sentient Labs \\
   \texttt{pramod@sentient.xyz} \\
}

\begin{document}

\maketitle

\begin{abstract}
  Law has long been a domain that has been popular in natural language processing (NLP) applications. Reasoning (ratiocination and the ability to make connections to precedent) is a core part of the practice of the law in the real world. Nevertheless, while multiple legal datasets exist, none have thus far focused specifically on reasoning tasks.  We focus on a specific aspect of the legal landscape by
  introducing a corporate governance reasoning benchmark (``CHANCERY") to test a model's ability to reason about whether executive/board/shareholder's proposed actions are consistent with corporate governance charters. This benchmark introduces a first-of-its-kind corporate governance reasoning test for language models – modeled after real world corporate governance law. The benchmark consists of a corporate charter (a set of governing covenants) and a proposal for executive action. The model’s task is one of binary classification: reason about whether the action is consistent with the rules contained within the charter. We create the benchmark following established principles of corporate governance - 24 concrete corporate governance principles established in \cite{gompers2003corporate} and 79 real life corporate charters selected to represent diverse industries from a total dataset of 10k real life corporate charters. Evaluations on state-of-the-art (SOTA) reasoning models confirm the difficulty of the benchmark, with models such as Claude 3.7 Sonnet and GPT-4o achieving 64.5\% and 75.2\%  accuracy respectively. Reasoning agents exhibit superior performance, with agents based on the ReAct and CodeAct frameworks scoring 76.1\% and 78.1\% respectively, further confirming the advanced legal reasoning capabilities required to score highly on the benchmark. We also conduct an analysis of the types of questions which current reasoning models struggle on, revealing insights into the legal reasoning capabilities of SOTA models.
\end{abstract}

\section{Introduction}

Legal reasoning represents one of the most advanced human cognitive abilities, requiring a deep understanding of domain knowledge and deductive inference to apply legal principles to a wide array of complex scenarios. It is also one of the most common forms of human cognitive reasoning, used on a daily basis in every jurisdiction around the world. As such, NLP applications have long focused on the legal domain, including popular tasks such as named-entity recognition, classification, and summarization \cite{quevedo2024,hachey2006,kanapala2019,chalkidis2020legal}. While the space of legal benchmarks is fairly saturated \cite{koreeda2021contractnli,guha2023legalbench,hendrycks2021cuad,wang2023maud}, few existing benchmarks test the advanced reasoning capabilities of ratiocination and analysis essential to many legal tasks. With the rise of advanced reasoning models such as OpenAI's o1/o3 models and DeepSeek's open-source R1 model, it has become increasingly necessary to develop benchmarks which test a model's ability to reason about complex legal texts. 

In this paper, we introduce CHANCERY - a novel dataset designed to bridge the gap between traditional legal benchmarks and the kind of advanced reasoning test required to evaluate the capabilities of current state-of-the-art models. The benchmark consists of 502 carefully crafted questions with manually designed prompts fed into GPT-4o spanning 79 distinct corporate charters from diverse industries. Our development process was guided by the seminal work of \cite{gompers2003corporate}, which identified 24 fundamental corporate governing principles from an analysis of 10,000 corporate charters. By systematically incorporating these principles into our benchmark along with real-life corporate charters, we ensure comprehensive coverage of the corporate governance landscape. 

\begin{figure}[!t]
    \centering
    \begin{subfigure}[b]{0.48\textwidth}
        \centering
        \begin{tcolorbox}[colback=gray!5!white, colframe=black, width=\textwidth, title=CHANCERY Benchmark, fontupper=\tiny, fontlower=\tiny]
        \tiny
        \textbf{Question:} A shareholder controlling 20\% of Southwest Airlines' stock has drafted a proposal to introduce cumulative voting. Is this proposal likely to succeed under the current charter rules?
        \vspace{0.5em}
        \textbf{Relevant Charter Clause (Excerpt):}
        \begin{tcolorbox}[colback=white, colframe=black!50, boxrule=0.5pt, sharp corners=south, fontupper=\tiny]
        \textbf{CHARTER – Article I:} Southwest Airlines Co., pursuant to the provisions of Article 4.07 of the Texas Business Corporation Act, hereby adopts Restated Articles of Incorporation which accurately copy the Articles of Incorporation and all amendments thereto that are in effect to date, and such Restated Articles of Incorporation contain no change in any provision thereof...
        \end{tcolorbox}
        \vspace{0.5em}
        \textbf{Answer:} \textit{No.}
        \vspace{0.5em}
        \textbf{Explanation:}
        Article Four explicitly prohibits cumulative voting. Any amendment to this provision would require compliance with Article Nine, which mandates an 80\% shareholder approval threshold for such changes, making it highly unlikely.
        \end{tcolorbox}
        \caption{Example from the CHANCERY benchmark.}
        \label{fig:charter-example}
    \end{subfigure}
    \hfill
    \begin{subfigure}[b]{0.5\textwidth}
        \centering
        \includegraphics[width=\textwidth]{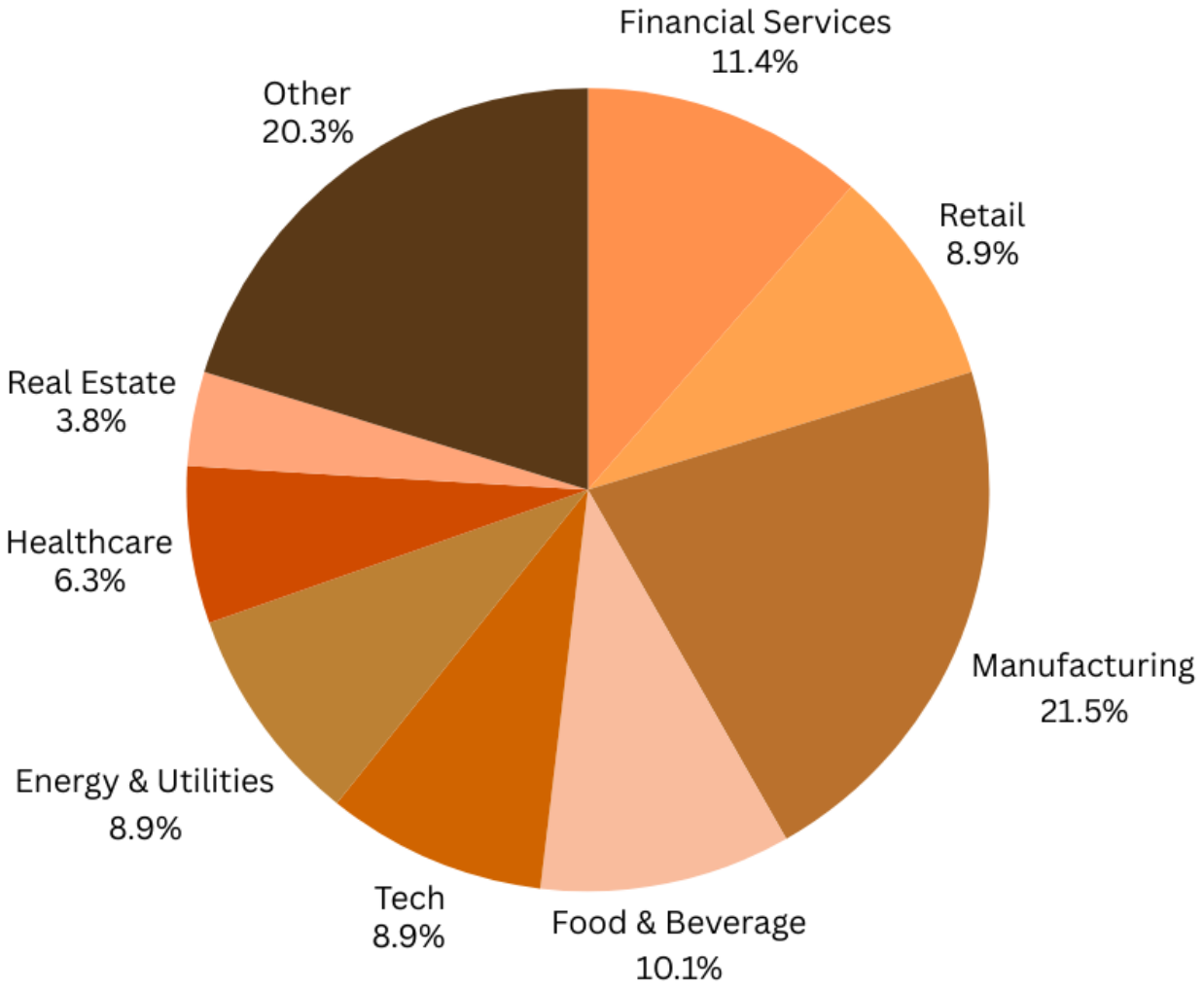}
        \caption{We include verticals from a diverse set of industries in our benchmark.}
        \label{fig:charter-industries}
    \end{subfigure}
    \caption{CHANCERY Benchmark example and industry distribution.}
    \label{fig:combined}
\end{figure}

The model's task is one of {\em binary classification}: determining whether an executive, board, or shareholder's proposed action is valid according to the provided charter; \ref{fig:charter-example} displays an example taken from the benchmark. Our dataset draws from real-world corporate charters and handcrafts questions to be challenging multi-step questions which require external knowledge, enhancing both its difficulty and comprehensiveness. Our evaluations  confirm the difficulty of the benchmark. State-of-the-art reasoning models struggle to achieve high performance, with Claude 3.7-Sonnet and DeepSeek-R1 scoring 64.5\% and 58.6\% accuracy respectively. These results underscore the substantial gap between current AI capabilities and the sophisticated reasoning required in binary classification tasks in real-world legal contexts. 

Beyond its immediate utility as an evaluation tool, the CHANCERY Benchmark contributes to a deeper understanding of language models' reasoning capabilities. Alongside our evaluations, we conducted a numerical analysis of DeepSeek-R1's performance across three separate verticals: (1) queries split by principles (2) queries split by single-hop vs.~multi-hop reasoning and (3) queries split by whether they require an extra search or not. We find that models struggle on particular corporate governance principles including Secret Ballots, Anti-greenmail, and Poison Pills as well as multi-hop queries and queries that require an additional search. Our results offer  insights into the strengths and weaknesses of current reasoning models' ability to reason about corporate governance and legal texts more broadly, serving as a  foundation for further work in the field.

\section{Related Work}
Recent advances in NLP have ushered in an array of datasets focused on the legal domain \cite{cui2023survey,xiao2018cail2018,zheng2021does,ma2021lecard}. Despite this, reasoning remains underrepresented in current benchmarks as compared to more traditional classification tasks \cite{wei2018empirical,shaheen2020large,de2020victor,undavia2018comparative,wan2019long}.  

\textit{LegalBench:} A foundational dataset in the legal reasoning domain, it consists of 162 tasks across 40 different contributors and is the largest collaborative effort for legal reasoning evaluation. The benchmark includes binary classification tasks (e.g Does the following qualify as hearsay?) and supervised learning tasks (e.g When does the Confrontation Clause apply?). However, despite its size, LegalBench notably misses out any reference to corporate governance \cite{guha2023legalbench}.

\textit{ContractNLI:} Another foundational benchmark focused on the legal reasoning domain, this tests whether models can classify contractual hypotheses as entailed by, contradicted, or unaddressed by the text of a supplied contract, while also identifying supporting evidence for their answers. While the task does test the reasoning abilities of language models, the authors do not focus on corporate governance \cite{koreeda2021contractnli}.

\textit{CUAD: Expert-Annotated Contract Review:} The CUAD dataset  \cite{hendrycks2021cuad} introduced 13k expert annotations across 41 clause types in 500+ contracts, achieving 72\% F1 with RoBERTa-large. While CUAD tests a model's ability to successfully identify clauses (e.g., "Governing Law" or "Perpetual License"), it does not test normative reasoning about whether detected clauses permit specific executive actions.

\textit{LexGLUE:} A legal benchmark which unified seven tasks including case outcome prediction (ECtHR) and statute entailment, with Legal-BERT achieving 71.2\% accuracy on case law entailment \cite{chalkidis2021lexglue}. However, its narrow focus on existing task formulations (e.g., binary statute support) fails to capture corporate governance’s  reasoning dimension- the need to balance strict charter language with  principles of fiduciary duty. 

\textit{Legal-BERT:} The LEGAL-BERT framework \cite{chalkidis2020legal} demonstrated that domain-specific pretraining on 12GB of legal corpora-including US case law (164k cases), EU legislation (116k documents), and SEC contracts (76k filings)-improves performance on legal NLP tasks by 5–12\% over generic BERT. However, its evaluation focused on token-level tasks (NER, text classification) rather than complex reasoning.








\section{Benchmark}

The CHANCERY benchmark consists of a binary classification task and was designed to be challenging for current SOTA reasoning models such as DeepSeek-R1 and Claude 3.7-Sonnet. 

\subsection{Benchmark Overview}
 The benchmark contains three fields: question, charter-id and answer. \textit{Question} is the executive/shareholder/board's proposal for an action. \textit{Charter-id} is the id of the charter taken from a dataset of 10k company charters. Charter-id corresponds to an id within the charter dataset, which is used to fetch the text of the company charter itself. \textit{Answer} is a yes or no binary target response which serves as the gold standard for a language model (i.e., whether the proposed action is consistent with the company charter). The benchmark contains 502 handcrafted questions which draw on a total of 79 distinct corporate charters. The ground truths were hand-labeled by the research team based on the questions produced. 

\subsection{Benchmark Diversity}

We enhanced the diversity of our benchmark in two key ways. First, we ensured that the questions included a comprehensive range of topics from corporate governance. We accomplished this by framing each executive/board/shareholder proposal in the benchmark to fit into 24 concrete corporate governance principles widely accepted in previous canonical work. Table~\ref{tab:model_performance} lists each of these principles and describes their function and purpose.  

Second, we made sure to include corporate governance charters from a broad range of different industry verticals. Figure~\ref{fig:charter-industries} displays the breakdown of charters included in the benchmark by industry. These included Manufacturing (17), Financial Services (9), Food \& Beverage (8), Retail (7), Tech (7),
Energy \& Utilities (7), Healthcare (5), Real Estate (3) and Other (16). 

\begin{table}[htbp]
    \centering
    \small 
    \begin{tabular}{|p{0.28\textwidth}|p{0.68\textwidth}|}
        \hline
        \textbf{Principle} & \textbf{Description} \\
        \hline
        Anti-greenmail & Prevention method against the agreement between a large shareholder and company, in which the shareholder sells their stock at premium price to the company and promises to refrain from influence on the company for some time. Anti-greenmail provisions usually force companies to make the same repurchase offer to every shareholder or to get majority shareholder approval. \\
        \hline
        Blank Check & Preferred stock with extensive rights that are issued by board without any shareholder approval. Mainly used as a defensive tactic against takeovers. \\
        \hline
        Bylaw/Charter Amendment Limitations & Restricts the shareholders' power to amend governance policies of the company. \\
        \hline
        Control-Share Cash-Out Laws & Allows shareholders to sell their stocks to a big shareholder based on the highest price of recently acquired shares. \\
        \hline
        Classified Board & Board structure where directors serve multi-year terms but with different end dates. This causes only a part of the board to be replaced each year making hostile takeovers harder. \\
        \hline
        Compensation Plans with Changes-in-Control & Compensation plans that enable participants to receive enhanced benefits in a case of change of control. \\
        \hline
        Director Indemnification Contracts & Contracts that guarantee specific directors, officers will be reimbursed for legal expenses and liabilities during their service time. \\
        \hline
        Control-Share Acquisition Laws & Limitations on the voting rights of an acquiring shareholder once their ownership passes a certain threshold unless getting approval from disinterested shareholders. \\
        \hline
        Cumulative Voting & A shareholder can pool their voting rights in any manner they want. This allows minor shareholders to elect directors by concentrating their votes. \\
        \hline
        Directors' Duties Provisions/Laws & Provisions that allow directors to consider non-shareholder constituencies such as employees, host communities and suppliers when considering a merger. \\
        \hline
        Fair-Price Provisions/Laws & Rules that limit the price difference a bidder pays in two-tier offers. This prevents the later sellers from earning less compared to early sellers. \\
        \hline
        Golden Parachutes & Severance agreements that grants monetary or non-monetary compensation for senior executives in an event like termination, demotion or resignation due to change of control. \\
        \hline
        Director Indemnification & A mechanism where a company protects directors from legal liabilities caused from a lawsuit related to their corporate rule. \\
        \hline
        Limitations of Director Liability & Charter Provisions or Amendments that limit directors' personal liability for duty of care (though not for loyalty, or intentional misconduct). \\
        \hline
        Pension Parachutes & Rule that prevents an acquirer from using the resources in the pension fund of target to finance the acquisition. \\
        \hline
        Poison Pills & Mechanism that equips the existing shareholders with extensive rights in a case of events like hostile takeover bid. This dilutes the hostile bidder's controlling power in the company. \\
        \hline
        Secret Ballot (Confidential Voting) & Shareholder voting remains confidential in order to prevent management exerting pressure on other shareholders. \\
        \hline
        Executive Severance Agreements & These agreements guarantee high-level executives of protecting their positions or some type of compensation (usually enhanced) that are not affected by change of control. \\
        \hline
        Silver Parachutes & Like golden parachutes where participants receive severance or benefits in case of change in control, however this is granted to non-executive employees. \\
        \hline
        Special Meeting Limitations & Rules that restrict shareholder's power in calling special meetings by usually enforcing higher shareholder support or completely eliminating special meetings. \\
        \hline
        Supermajority Requirements & Provisions that require more than a simple majority to approve significant actions like mergers. \\
        \hline
        Acquisition Laws & Laws that regulate actions like mergers, takeovers and tender offers. \\
        \hline
        Unequal Voting Rights & Limits the voting rights of some shareholders and increases for others per share. It may be decided based on time-phases where longer shareholders have more voting power or substantial based where big shareholders have less voting power per share. \\
        \hline
        Limitations on Action by Written Consent & Forbids or limits the allowed shareholder actions without formal meetings to delay hostile efforts. \\
        \hline
    \end{tabular}
    \vspace{10pt} 
    \caption{The 24 Corporate Governance Principles Used in Our Benchmark}
    \label{tab:governance_principles}
\end{table}

\subsection{Benchmark Creation Process}
Our starting point is the now-classical work in \cite{gompers2003corporate}, which curated 10,000 real-life corporate charters and identified 24 ``governance principles'' that define and guide executive actions.
\begin{itemize}
\item For instance, the curated corporate charters included those from Southwest Airlines, Nordstrom, Heritage Financial Corporation, etc.
\item For instance, the identified corporate governance principles included Golden Parachutes, cumulative voting, Poison Pills, acquisition laws, etc.
\end{itemize}
We constructed our dataset through the following process:
\begin{enumerate}
\item We find that each actual corporate charter is a good candidate and hence chose a subset of the 10,000 charters at random in constructing our dataset (79 charters in particular).
\item We find that \textit{each} of the 24 different principles represents an important and different aspect of corporate governance and hence used each of the principles as a basis for creating our dataset.

\item The crucial part of the dataset creation process involved inventing an action that is inspired by one of the 24 guiding principles and then reasoning ourselves about whether such an action is consistent with a specific charter. We took generous assistance of existing AI models to invent such actions (e.g., prompting GPT-4o to create a hypothetical scenario using the charter and one or more of the 24 corporate governance principles). This process involved substantial iteration across expert human labor and experimentation with existing AI models.

\item The gold standard for the outcome of the binary classification is the decision carefully made by a human (one of us), together with a paragraph explaining the decision process and the outcome (mirroring the ``judgement'' from the bench of a court of law). Since the task is fairly sharp and logically clear, we did not find any discrepancy among the decisions (our explanation paragraphs) of the different humans working on this dataset.
\end{enumerate}
All the questions are alluding to hypothetical scenarios that are related to a charter. The difficulty of the dataset was enhanced in several key ways:
\begin{enumerate}
\item One of the ways that we used to increase the difficulty was by giving contradictive, and confusing information regarding the scenarios. This not only made the questions more complicated but also a lot longer.
\item We also made sure to use multiple sections of a charter and components of general law (such as Delaware General Corporation Law (DGCL)) in each question. After the model has investigated the scenario, it should have a general idea of what it needs to search.
\end{enumerate}
By using a combination of charter and general law, we ascertained that the model needs to understand both charter and general law completely.

Our benchmark and associated evaluation code are provided in the following anonymous GitHub link: \hyperlink{https://anonymous.4open.science/r/chancery-benchmark-775D/README.md}{https://anonymous.4open.science/r/chancery-benchmark-775D/README.md}.


\begin{table}[!t]
    \centering
    \begin{tabular}{lccc}
        \toprule
        \textbf{Models} & \textbf{Accuracy}  \\
        \midrule
        QwQ-32B & 55.4\% \\
        DeepSeek-R1 & 58.6\% \\
        Claude 3.7-Sonnet & 64.5\% \\
        Llama3.1-70B & 72.7\% \\
        GPT-4o & \textbf{75.2\%}\\
        \midrule
        \textbf{Reasoning Agents} \\
        \midrule
        ReAct agent & 76.1\%  \\
        CodeAct agent & \textbf{78.1\%} \\
        \bottomrule
    \end{tabular}
    \vspace{10pt} 
    \caption{Performance comparison of SOTA reasoning models and ReAct and CodeAct agents on the CHANCERY benchmark. Our evaluations demonstrate that even advanced reasoning models struggle to achieve a high accuracy on the CHANCERY benchmark, with the highest score achieved by our CodeAct agent.}
    \label{tab:model_performance}
\end{table}
\section{Experiments}
The difficulty of our dataset is reflected in the results obtained by running evaluations of top reasoning and non-reasoning models. 

\subsection{Model Evaluations}

Table~\ref{tab:model_performance} summarizes the performance of state-of-the-art (SOTA) language models on the CHANCERY benchmark. The benchmark reveals a significant gap between the performance of SOTA models and reasoning agents, underscoring the difficulty of the task and its heavy dependence on a model's specialized legal reasoning capabilities. 


General-purpose language models such as GPT-4o and Llama3.3-70B achieve the highest accuracies of 75.2\% and 72.7\% respectively, outperforming reasoning language models such as DeepSeek-R1 (58.7\%) and Claude 3.7-Sonnet (64.5\%). However, the results differed depending on the system prompt fed to the model. Without specialized prompts requiring the models to output only a single, binary yes or no answer, the performance of GPT-4o and Llama3.3-70B dropped significantly to 29.1\% and 22.1\% respectively. 

Overall, the accuracies achieved by the models we evaluated were relatively low. Even SOTA reasoning models like QwQ-32B (55.4\%) scored barely higher than the expected performance of a random coin flip. This indicates that advances in reasoning or model size alone are not sufficient for legal reasoning tasks such as the ones present in our benchmark; instead, our benchmark reveals that alignment with legal domain knowledge and the ability to perform multi-step deductive reasoning based on canonical legal texts (e.g., corporate charters) appears to be crucial to achieving high performance on legal reasoning tasks. 

Due to the nature of the reasoning involved in CHANCERY -- the multi-step aspect and need for tool-calling (e.g, search for legal precedent)  -- we hypothesize that SOTA agentic frameworks are better equipped to solve legal reasoning problems. We constructed two reasoning agents based on recent work which reported SOTA performance on the FRAMES reasoning benchmark \cite{alzubi2025open}. The first is an agent based on the Reasoning-and-Action (ReAct) agentic framework \cite{yao2023react}, and the second is an agent based on the CodeAct framework (which transforms text into code which the model can more easily interpret) \cite{li2023chain}. These agents outperform all tested reasoning models, scoring 76.1\% and 78.1\% respectively, confirming our hypothesis. However, they still score far less than would be expected from high legal reasoning capabilities, providing further evidence of the difficulty of our benchmark.

\begin{table}[htbp]
\centering
\caption{Accuracy on CHANCERY broken down by DeepSeek-R1's performance on each individual corporate governance principle. This analysis reveals that principles such as Anti-greenmail, secret ballots and poison pills are significantly harder for SOTA LLMs.}
\vspace{2mm}
\begin{tabular}{lr}
\toprule
\textbf{Corporate Governance Principle} & \textbf{Accuracy (\%)} \\
\midrule
Anti-greenmail & 27.6 \\
Secret ballots & 33.3 \\
Poison pills & 41.9 \\
Pension parachutes & 42.3 \\
Golden parachutes & 45.9 \\
Executive severance agreements & 47.0 \\
Blank check & 48.9 \\
Compensation plans with changes-in-control & 48.1 \\
Special meeting limitations & 49.8 \\
Control share acquisition laws & 50.7 \\
Limitations on action by written consent & 51.8 \\
Business combination laws & 52.0 \\
Classified board & 52.8 \\
Directors duties provisions/laws & 53.2 \\
Acquisition laws & 53.4 \\
Limitations of director liability & 53.6 \\
Silver parachutes & 53.3 \\
Director indemnification & 54.1 \\
Super-majority requirements  & 55.2 \\
Unequal voting rights & 55.7 \\
Director indemnification contracts & 56.4 \\
Fair-price provisions/laws & 57.7 \\
Bylaw and charter amendment limitations & 58.1 \\
Cumulative voting & 59.0 \\
Control-share cash out laws & 63.6 \\
\bottomrule
\end{tabular}
\label{tab:principles_performance}
\end{table}
\subsection{Benchmark Analysis}

By splitting the benchmark across three verticals: (1) principles (2) single-hop vs multi-hop queries and (3) queries that require an extra search vs no extra search, we find that current SOTA models perform better or worse depending on the type of query. Table 3 displays the accuracy achieved by DeepSeek-R1 on each  corporate governance principle. Anti-greenmail (27.6\%), Secret Ballots (33.3\%) and Poison Pills (41.9\%) were the most challenging principles while Bylaw and Charter Amendment Limitations (58.1\%), Cumulative Voting (59.0\%), and Control-Share Cash Out Laws (63.6\%) were the easiest. Additionally, we find that DeepSeek-R1 performed significantly worse when the questions were multi-hop (58.1\%) compared to single-hop (73.3\%). DeepSeek also performed worse when it required an additional search to cross reference existing legislation with the provided governance charter, scoring 58.0\% compared to 70.7\% when no cross-referencing was necessary. 

\begin{figure}[h!]
    \centering
    \begin{lstlisting}[basicstyle=\ttfamily\tiny, frame=single]
A question from CHANCERY: 
amid heightened market volatility and an increase in shareholder activism, the board of directors of ameron international corporation proposes the enhanced stability and governance initiative (esgi), a comprehensive restructuring plan aimed at reinforcing long-term corporate stability while mitigating the risks associated with external hostile takeovers and rapid market driven leadership changes. the esgi comprises the following key provisions:1. supermajority voting for governance amendments: any future amendments to the corporate charter, especially those concerning director elections, poison pill provisions, or board compensation structures, will require an affirmative vote of at least 80% of outstanding voting stock, in accordance with article fifth.2. expansion of director indemnification rights: the board proposes an amendment ensuring that all directors, regardless of tenure, receive full indemnification against legal claims, provided that their actions do not constitute fraud, willful misconduct, or gross negligence. this is aligned with the protections granted under article tenth(2) and further reinforced by delaware corporate law protections against director liability limitations.3. antigreenmail protections: to prevent ameron from being forced into stock repurchases at inflated premiums by activist investors, a clause is introduced restricting the corporation from purchasing stock from a substantial shareholder at a price above fair market value unless approved by two-thirds of the board of directors.4. director tenure protection: directors will continue serving in staggered terms as defined under article sixth, ensuring that no more than one-third of the board is up for election in any given year, preventing abrupt leadership changes.5. no-action by written consent: in accordance with article eighth, all shareholder actions must be conducted at a formal meeting, eliminating the possibility of hostile investors rapidly accumulating shares and pushing through governance changes without due process......can ameron international corporation legally enforce the enhanced stability and governance initiative (esgi)?

Correct Answer: 
   Yes. The ESGI aligns with Ameron International Corporation's existing charter provisions and Delaware corporate law. The supermajority amendment requirement is explicitly supported by Article SEVENTH, which already mandates an 80% threshold for critical governance changes. The director indemnification expansion does not violate Delaware law, as it does not indemnify against fraud, intentional misconduct, or gross negligence. Furthermore, Article SIXTH explicitly allows staggered board terms, meaning the proposal's Director Tenure Protection is entirely within Ameron's legal rights. The Anti-Greenmail Provision is a common anti-takeover mechanism that has been upheld in Delaware courts, and the No-Action by Written Consent Clause is directly supported by Article EIGHTH. As such, the ESGI is legally enforceable.
 
DeepSeek-R1 response: 
Okay, let's try to figure out whether Ameron International can legally enforce the ESGI. The question is about corporate law, specifically Delaware law since the company is incorporated there. The key here is to check if the proposed provisions in the ESGI are allowed under the company's Restated Certificate of Incorporation (the charter) and Delaware General Corporation Law (DGCL)....The existing amendment process for the charter is in Article ELEVENTH, which says that amendments to the charter require the vote prescribed by law (which is majority unless the charter says otherwise), except for the listed articles (FIFTH, SIXTH, etc.), which require 80%. So if the ESGI is adding a new article or amending an existing article not in the protected list, then a majority vote would suffice. If they are amending one of the protected articles, they need 80%. For example, if the supermajority requirement for charter amendments is added to Article FIFTH (which is about bylaws), then they would need 80% to amend it.

[OMITTED]

   ANSWER
   NO
   
    \end{lstlisting}
    \caption{An example question and DeepSeek-R1 answer from the CHANCERY benchmark based on governance principles used}
    \label{fig:principle_example}
\end{figure}

The example displayed in \ref{fig:principle_example} illustrates a question DeepSeek-R1 struggled with due to the presence of a more challenging corporate governance principle. The question draws on five different corporate governance principles: Supermajority Requirements, Director Indemnification, Anti-Greenmail, Limitations on Action by Written Consent and Classified Boards. DeepSeek-R1 fails to synthesize the information from the charter with the ESGI proposal but rather tries to reason about each of the five principles separately, causing it to miss connections between the ESGI and what was stated in the charter. Its pitfall was in its inability to understand the Anti-greenmail principle (the hardest of the principles according to \ref{tab:principles_performance}). The charter doesn’t explicitly state anything related to Anti-Greenmail and thus the model would need to possess an a priori understanding of the principle or otherwise conduct a web search to gather information about it. The model presumably disregarded this principle as it was not directly stated in the company’s charter. It failed to recognize the importance of Anti-greenmail while exaggerating the proposed power of the ESGI based on other principles. Because of this over-complication of the arguments, it is highly probable that the model said NO to the question in order to give a conservative answer, when the gold standard was an affirmative YES. Further examples illustrating multi-step queries and queries that require an additional search are provided in the Appendix.

\section{Discussion}

The rise of highly capable reasoning models such as Claude 3.7-Sonnet and DeepSeek-R1 have raised the bar for NLP benchmarks. One capability which has remained conspicuosly  underrepresented is {\em legal reasoning}, despite the fact that it represents one of the most advanced cognitive abilities humans possess. In this work, we introduced a novel legal reasoning benchmark we call  “CHANCERY”, to bridge the gap between traditional NLP legal benchmarks and the type of advanced reasoning benchmark required to evaluate current state-of-the-art reasoning models. Evaluations we performed confirm the difficulty of the benchmark, with reasoning agents based on the ReAct and CodeAct frameworks exhibiting the best performance. 

While our benchmark is a first-of-its-kind reasoning dataset for corporate governance, there are three key ways in which it could be improved. First, the benchmark could be expanded to increase the number and diversity of the included charters. In our work, we only use 79 charters from the 10k+ provided in \cite{gompers2003corporate}. We could expand our benchmark by drawing on a greater number of charters from the dataset and considering other high-quality sources of governance charters. While we tried to represent a breadth of industry verticals in the charters we included, the diversity of industries could be further expanded upon and explore the industry-specific nuances of the application of the law. 

Second, the benchmark is currently highly
US-centric. All 79 charters are drawn solely from companies that are headquartered in US states. An expanded dataset covering other jurisdictions  including cross-country law (a specialty in the real world) could be used to explore the generalizability (or its absence) of reasoning across jurisdictions. Third, the dataset could be expanded to encompass other forms of governance. While there is a long history of traditional corporate governance, new forms of company governance have emerged to govern industries such as the cryptocurrency and blockchain industry (under the rubric of DAO – distributed autonomous organization). The benchmark could be improved by including these charters, typically far more nebulous and well defined than the Delaware corporate charters we studied in the current work. 

In summation, our results offer  insights into the legal reasoning capabilities of current state-of-the-art reasoning models. Building on our work with CodeAct and ReAct agents, future directions might test the performance of custom-built legal reasoning agents on our benchmark. The CHANCERY benchmark serves as a strong foundation for testing the legal reasoning abilities of current reasoning models, and we invite others to build upon this basis in future work.

\section*{Acknowledgements}

We thank Sreeram Kannan for insightful conversations and the EigenLayer team for developing the Verifiable Cloud framework, that have significantly motivated the ideas presented in this work.

\clearpage
\bibliographystyle{plainnat}
\bibliography{sample}

\appendix

\section{DeepSeek-R1 Example Answers}
\begin{lstlisting}[basicstyle=\ttfamily\tiny, frame=single]
A question from CHANCERY: 
in response to an emerging financial crisis triggered by an unexpected collapse in regional banking markets, north fork bancorporation, inc. finds itself under immense pressure from both regulatory bodies and institutional investors to shore up its financial stability while maintaining stockholder confidence. to address these concerns, the board of directors implements the **liquidity and market stability initiative (lmsi)** a multi-faceted restructuring plan designed to bolster the company's capital reserves, restructure governance mechanisms, and introduce defensive measures against speculative short-selling attacks. the lmsi consists of the following five major provisions: ........given these competing arguments, can the board legally proceed with the liquidity and market stability initiative?

Correct Answer: 
Yes. The Board of North Fork Bancorporation, Inc. can legally proceed with the **Liquidity and Market Stability Initiative (LMSI)** under the authority granted by the company's charter....while the LMSI may be challenged as aggressive and shareholder-unfriendly, it **does not violate the company's charter**, allowing the Board to implement the plan.
 
DeepSeek-R1 response: 
Okay, let's try to figure out whether the Board of North Fork Bancorporation can legally proceed with their LMSI plan. There are five provisions being challenged, each with arguments from both the Board and the activist investors. I need to go through each one step by step, check the relevant parts of the Certificate of Incorporation provided, and see which side has the stronger legal standing. 


[OMITTED]
   ANSWER
   NO
   
    \end{lstlisting}

This question is a multi-step question. Liquidity and Market Stability Initiative (LMSI) combines five major governance policies. The question requires the model to analyze multiple charter sections (FOURTH (b), (e), (f), (g); SIXTH; SEVENTH) and to assess if each of the provisions is legal with regard to charter and Delaware Law. For instance,  VRRM depends on how voting reclassification is applied, while AHTS forces the model to investigate Unocal standards with the board's charter authority. Overall, all the propositions that the LMSI makes have the same degree of relation to each other and, therefore, requires the model to reason through multiple aspects of the question and charter to answer the question. The model gave a false answer as it failed to interpret the charter provisions extensively and Delaware Corporate Law regarding board discretion during financial crises. For example, the model deemed VRRM as illegal as it assumed that the provision was going to modify the voting rights of existing common shares without refuting the board's claim that this move is justified under Article FOURTH,Section(b)(g). Similarly, model also refused the Director Tenure Extension Clause, claiming that it was a violation to shareholder election rights. However, it failed to recognize Article SIXTH that allows these types of extensions and Delaware  Law that allows flexibility for boards with good faith to continue stability during crises. Overall, the main issue with the model was that it directly came to a conclusion regarding questionable provisions when a clause related to that provision was not explicitly stated in the charter. This led the model to disregard subtle explanations in the charter that allowed discretionary powers to board especially during problematic times like Article FOURTH on capital management and dividends or Article SEVENTH on bylaw control. 

\begin{lstlisting}[basicstyle=\ttfamily\tiny, frame=single]
A question from CHANCERY: 
after the compensation committee of Red Robin Gourmet Burgers, Inc. authorized a multi-layered executive severance framework that blended traditional cash-based golden parachute triggers with equity vesting acceleration clauses tied to 'perceived instability events,' an investor coalition led by a syndicate of institutional governance monitors and pension advisory funds challenged the legality of the plan on the basis that it created structurally irreversible payouts with no meaningful review mechanisms. The compensation plan, approved by a 5-3 board vote, granted all C-level executives an immediate cash severance package equal to three times base salary and full vesting of all performance-based restricted stock units (RSUs), contingent on either (i) termination without cause or (ii) any board-approved declaration of a 'potential control-related risk event' that would trigger the payout schedule irrespective of actual acquisition activity. Concurrently, the plan introduced a 'double-qualifier' pension parachute clause, ....the court was asked to determine whether the board's adoption of a severance and pension parachute plan that provides payouts based on internal governance stress events, without shareholder approval or external triggering conditions, was consistent with the authority granted under the charter and Delaware corporate law.

Correct Answer: 
No. The executive severance and pension parachute plan adopted by the board of Red Robin Gourmet Burgers, Inc., as structured, exceeds the permissible boundaries of board authority under both the company's Amended and Restated Certificate of Incorporation and Delaware corporate law. Article FIFTH(A) of the charter indeed grants the board general managerial authority, stating that 'the business and affairs of the Corporation shall be managed by or under the direction of the board of directors.' However, this delegation is not absolute and must be read in conjunction with fiduciary duties and shareholder governance rights codified under Delaware law. The severance plan in question.....undermines the power of shareholders to hold the board accountable-particularly the rights to remove directors for cause under Article SIXTH(D) and to influence corporate governance direction through ordinary elections.....While the charter does not explicitly bar such plans, its silence does not confer a blank check. Fiduciary constraints and the interlocking structure of Articles FIFTH and SIXTH establish a governance regime in which executive exit packages must remain subordinate to the balance of shareholder authority and director accountability. As constructed, the plan creates economic deterrents to governance reform and rewards strategic alignment with incumbent leadership, violating the spirit and operational bounds of both charter structure and Delaware common law. Therefore, the severance plan is inconsistent with Red Robin's charter authority and Delaware fiduciary standards.
 
DeepSeek-R1 response: 
Okay, let's try to work through this problem. So, the question is about whether the board of Red Robin Gourmet Burgers, Inc. acted within their authority under the company's charter and Delaware law when they adopted this severance and pension parachute plan. The plan gives executives payouts based on internal governance stress events without needing shareholder approval or external triggers. The investors are challenging this, saying it's like a poison pill that entrenches executives and violates the charter and Delaware law.

[OMITTED]
   ANSWER
   YES
   
    \end{lstlisting}
    \label{fig:source_exmaple}

This question is an external question as the question depends on principles like board entrenchment, fiduciary duties and anti-takeover doctrine which are not explicitly stated in charter so the model needs to refer to external sources which leads the model to extensively use Delaware law (Unocal, Blasius etc.). We can also see usage of phrases like “Delaware courts have consistently held…” and “fiduciary constraints…”  which are indicators of usual practices of company governance that count as external information. The model incorrectly concluded that the severance plan was invalid. However under Article FIFTH(a), the board has board discretion to manage corporate affairs, which includes compensation, and there is no explicit restriction to what severance plan proposes. Model overweights external fiduciary law (e.g. Unocal, Blasius) without properly analyzing how it is relevant to charter. It directly assumes that discretionary severance triggers constitute entrenchment without explaining how they harm shareholder voting or removal rights under Article SIXTH(D). In general, the model mostly focused on what Delaware general law stated regarding the entrenchment which resulted in the model to underweight the charter which didn’t prohibit the plan. Leaning towards general fiduciary law, caused a model to miss that courts often transfer the power to boards regarding policies about executive compensation.

\section{Model Prompts}
\label{appendix}
We used a strict binary classification prompt to evaluate model performance on charter question-answering. The following prompt template was used for all experiments:
\begin{lstlisting}[breaklines=true, frame=single, basicstyle=\small\ttfamily]
Based on the charter text below, please answer the following question with ONLY 'Yes' or 'No'.
Question: {question}
Charter Text: {charter_text}
Your response must be EXACTLY ONE WORD: either 'Yes' or 'No'. No additional text or explanation is allowed.
\end{lstlisting}
This strict formatting was enforced to improve performance of model responses across all experimental conditions. The models were instructed to respond with only a single word to enable direct comparison with ground truth labels. This is the prompt used to achieve higher performance on GPT-4o and Llama3.3-70B.

\end{document}